\documentclass{article}

\usepackage{amsmath}
\usepackage[preprint]{neurips_2026}
\makeatletter
\renewcommand{\@notice}{}
\makeatother
\usepackage[utf8]{inputenc} 
\usepackage[T1]{fontenc}    
\usepackage{hyperref}       
\usepackage{url}            
\usepackage{booktabs}       
\usepackage{amsfonts}       
\usepackage{amssymb}        
\usepackage{nicefrac}       
\usepackage{microtype}      
\usepackage{graphicx}       
\usepackage{multirow}       
\usepackage{xcolor}         
\usepackage{amsthm}
\usepackage{placeins}
\hypersetup{hidelinks}

\title{Valid Early Detection of Memorization in Diffusion Models}
\workshoptitle{E-Values: From Statistics to ML}

\author{%
  \begin{tabular}{cccc}
    Yixiao Jing & Chia-Yu Hsu & Milad Hoseinpour & Zekai Zhang \\
    \texttt{yxjing@umich.edu} & \texttt{chiayuh@umich.edu} &
    \texttt{miladh@umich.edu} & \texttt{zzekai@umich.edu}
  \end{tabular}\\[0.5em]
  Department of Electrical Engineering and Computer Science\\
  University of Michigan\\
  Ann Arbor, MI, USA
}

\begin{document}

\maketitle

\begin{abstract}
    Diffusion models can sometimes reproduce examples from their training data, a phenomenon known as memorization. Existing memorization detection methods for unconditional diffusion models typically operate at a pre-selected timestep and therefore cannot adapt the detection time to the evolving denoising trajectory. Sequential monitoring enables adaptive detection but repeated testing across timesteps complicates trajectory-level false-alarm control. We study two ways of combining split conformal calibration with sequential monitoring for prompt-free memorization detection: per-step Bonferroni calibration and direct trajectory-level calibration of the maximum memorization score. The former controls repeated testing by splitting the false-alarm budget across timesteps, whereas the latter calibrates the event of ever crossing a threshold directly through the trajectory-level maximum. We evaluate this formulation on an unconditional CIFAR-10 EDM using two complementary per-step statistics based on geometry and representation spikiness. Experiments show that sequential monitoring improves the power--timing tradeoff over fixed-time detection. Compared with Bonferroni calibration, trajectory-level calibration achieves comparable detection power while using the false-alarm budget less conservatively and enabling earlier detection at higher noise levels.
\end{abstract}
\begin{figure}[!ht]
  \centering
  \includegraphics[width=0.8\linewidth]{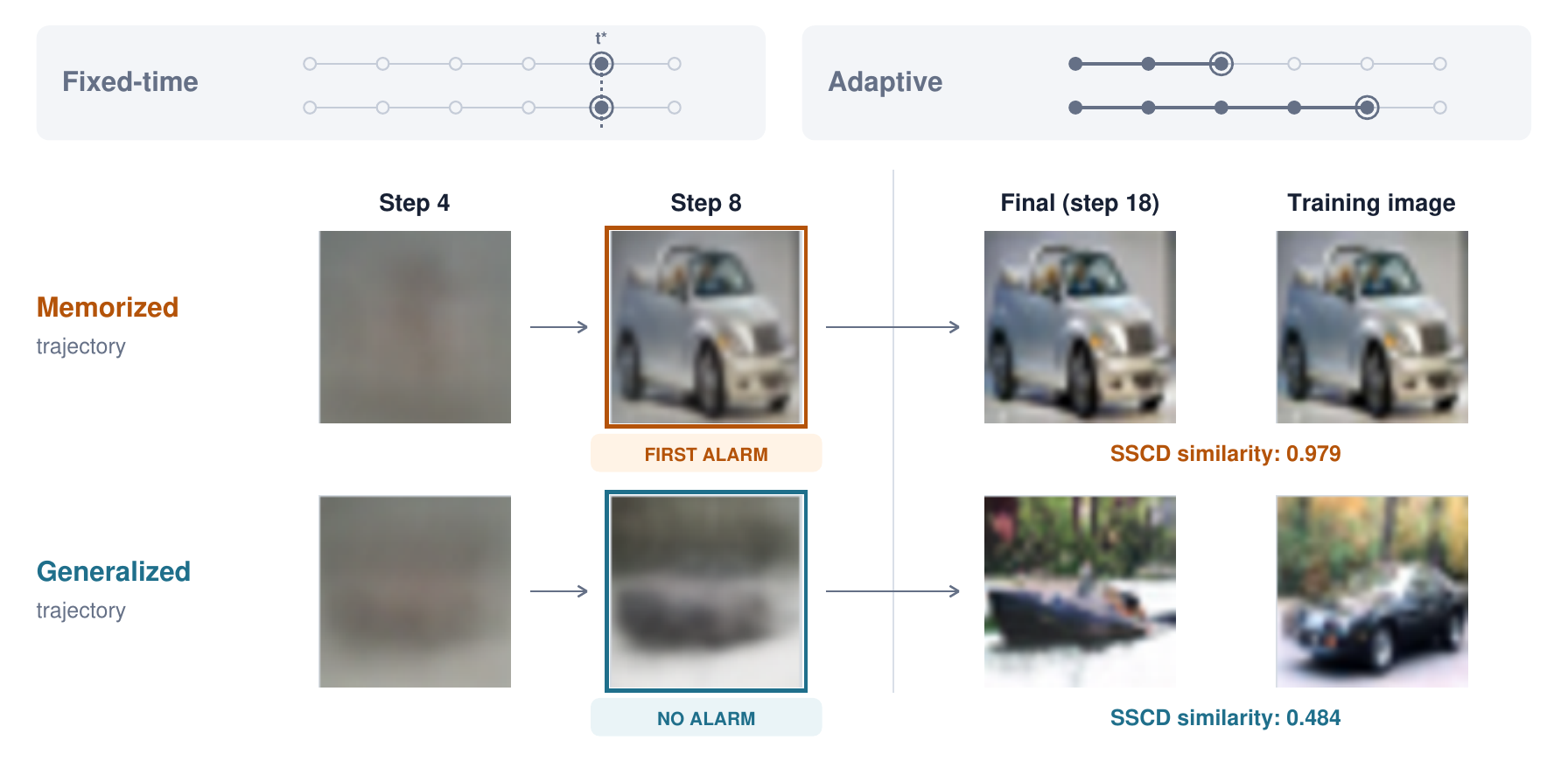}
  \begingroup
  \color{black}
  \caption{\textbf{Early warning before generation finishes.} Top shows schematic comparison of fixed-time checks at a preselected step and adaptive alarms at trajectory-dependent steps. Intermediate panels show the model's denoised predictions at selected sampling steps. The memorized trajectory triggers its first alarm at step 8 of 18, while the generalized trajectory triggers no alarm. Final outputs are shown beside their nearest training images, with SSCD similarity scores.}
  \label{fig:teaser}
  \endgroup
\end{figure}

\section{Introduction}
Diffusion models have demonstrated remarkable generalization capabilities, enabling the generation of diverse and novel images \citep{kadkhodaie2024generalization, zhang2024emergence}. However, a growing body of research has shown that these models may also
reproduce samples from their training data, a phenomenon known as memorization \citep{carlini2023extracting, somepalli2023diffusion,
somepalli2023understanding}, which substantially reduces the novelty and privacy of the generated content.

Motivated by this concern, recent work has investigated the mechanisms and detection of memorization in diffusion models. One line of work focuses on understanding when and how memorization emerges during the denoising process. Research suggests that memorization is most pronounced during an intermediate stage of denoising \citep{dodson2026two, kaiser2026diffusion}, while its effects become apparent near the end of the generation process \citep{biroli2024dynamical}. Although this line of work does not provide an explicit detection procedure, it offers important insights into when memorization-related signals emerge during denoising and suggests that informative statistics may emerge before its effects become apparent.

The second line of work focuses on distinguishing trajectories that ultimately lead to memorization. Existing detection methods fall into two groups. The first group is prompt-based, which exploits text-conditioning statistics and can enable early detection \citep{wen2024detecting, ren2024unveiling}. However, these methods are not applicable to unconditional diffusion. The second group is prompt-free, relying on geometric or representation-based statistics to detect memorization \citep{ross2025geometric, zhang2026balanced}. Nevertheless, these approaches require the selection of a fixed noise level or timestep and are typically applicable only relatively late in the denoising process, while early detection can further enable inference-time mitigation before generation is completed \citep{wen2024detecting}. \citet{chen2024towards} instead monitor
memorization similarity throughout denoising with time-dependent
thresholds, but do not provide statistical control of trajectory-level
false alarms.

This leaves a gap for prompt-free detection: fixed-time procedures cannot adapt the detection time to individual trajectories, whereas sequential monitoring enables such adaptivity but introduces repeated-testing effects. Figure~\ref{fig:teaser} illustrates our goal of detecting memorizing trajectories during sampling without falsely alarming generalized trajectories. We study how split conformal calibration \citep{vovk2005algorithmic, angelopoulos2021gentle} can be combined with sequential monitoring in this setting. We compare three strategies: fixed-time calibration, per-step Bonferroni calibration for sequential monitoring, and direct trajectory-level calibration of the maximum memorization score. The trajectory-level strategy calibrates the maximum score over the trajectory, treating the entire monitoring event as a single scalar statistic. We establish finite-sample false-alarm control for all three strategies and show empirically that sequential monitoring improves the power--timing tradeoff over fixed-time detection, while trajectory-level calibration is less conservative and enables earlier detection than per-step Bonferroni calibration.

\section{Problem Formulation}

\subsection{Diffusion Trajectories and Memorization}
\label{sec:memorization}

A diffusion model generates a sample by progressively denoising an initial noise sample. To facilitate sequential testing, we index the reverse process in increasing order and define a denoising trajectory as follows
\begin{equation}
    x_{1:T} = (x_1, x_2, \ldots, x_T),
\end{equation}

where \(x_1\) is the initial noisy
sample and \(x_T\) is the final generated image. Throughout, we use a deterministic \(T=18\)-step EDM sampler whose noise levels decrease from \(\sigma_1=80.0\) to \(\sigma_T=0.002\), where \(\sigma_t\) is the EDM noise level at step \(t\) and a higher \(\sigma_t\) corresponds to an earlier, less-denoised stage of generation (Appendix~\ref{appendix:exp-setup}).

We cast the detection of memorization as a sequential testing problem
\begin{equation}
    H_0: x_{1:T} \sim P_{\mathrm{gen}},
    \qquad
    H_1: x_{1:T} \sim P_{\mathrm{mem}},
\end{equation}

where \(P_{\mathrm{gen}}\) and \(P_{\mathrm{mem}}\) denote the distributions of generalized and memorized trajectories, respectively; we write \(P_{H_0}\) and \(P_{H_1}\) for probabilities under these two distributions.
Operationally, we label a trajectory by the largest cosine similarity between \(x_T\) and any training image under a pretrained SSCD encoder \citep{pizzi2022sscd} with generalized below \(0.5\) and memorized above \(0.9\). The intermediate band is excluded and no guarantee is claimed for it (detailed in Appendix~\ref{appendix:exp-setup}).

\subsection{Detection Statistics and Memorization Scoring}
\label{sec:statistics-scoring}

We use two per-step scalar statistics that capture distinct mechanisms, showing that the framework is not tied to a specific statistic. We denote the statistic at timestep \(t\) generically by \(M_t\).

The first statistic, trF, is motivated by the density-sharpness characterization of memorization in \citet{jeon2025understanding}. It is the trace of the Jacobian of a network function \(F_\theta\) with respect to its input \(u_t\):
\begin{equation}
    M_t^{\mathrm{trF}} = \operatorname{tr}\!\left(\frac{\partial F_\theta(u_t)}{\partial u_t}\right),
    \label{eq:trf-statistic}
\end{equation}
where \(F_\theta\) is the raw EDM network output and \(u_t\) its preconditioned input.

The second statistic, StD, measures representation spikiness \citep{zhang2026balanced} through the channel-wise standard deviation after spatial global-max pooling:
\begin{equation}
    M_t^{\mathrm{StD}} = \operatorname{Std}(\text{GlobalMaxPool}(h_t)),
    \label{eq:std-statistic}
\end{equation}
where \(h_t\) denotes an intermediate network representation at timestep \(t\).

Network choices, input preconditioning, trace estimation, and further implementation details are provided in Appendix~\ref{appendix:statistics}.

To train and calibrate the scoring model, we split the labeled trajectories \((x_{1:T}^{(i)},L^{(i)})\), where \(L^{(i)}=0\) for generalized and \(L^{(i)}=1\) for memorized trajectories, at the trajectory level into a fitting set \(\mathcal{D}_{\mathrm{fit}}\), a validation set \(\mathcal{D}_{\mathrm{val}}\), and a test set \(\mathcal{D}_{\mathrm{test}}\). Conformal thresholds are computed from a null calibration set \(\mathcal{D}_{\mathrm{cal}}\) of held-out generalized trajectories in \(\mathcal{D}_{\mathrm{test}}\); false alarms and detection power are evaluated on the remaining generalized and memorized test trajectories, respectively. The detailed splits are provided in Appendix~\ref{appendix:exp-setup}.

Given the statistic history \(M_{1:t}=(M_1,\ldots,M_t)\) up to timestep \(t\), we train a causal scoring model
\begin{equation}
    \hat q_t = g(M_{1:t}),
\end{equation}
where \(g\) is implemented as a causal Gated Recurrent Unit (GRU) \citep{cho2014learning} and \(\hat q_t\) is a memorization score based only on information available up to timestep \(t\).

At every timestep, $g$ is trained with binary cross-entropy (BCE) against the final label $L$, so that $\hat q_t$ estimates $q^\star_t=P(L=1\mid M_{1:t})$, the probability that the trajectory ends memorized given the statistics observed so far.
\section{Valid Sequential Detection}

\subsection{Calibration Strategies}
\label{sec:calibration-strategies}

For a target trajectory-level false-alarm level \(\alpha\), we consider fixed-time calibration, per-step Bonferroni calibration, and trajectory-level max calibration. The key idea of the trajectory-level
strategy is to calibrate the event of ever crossing a common threshold as a
single trajectory-level statistic:
\begin{equation}
    \{\exists\, t\le T:\hat q_t>c\}
    =
    \{\max_{t=1,\ldots,T}\hat q_t>c\}.
\end{equation}
Let \(n_{\mathrm{cal}}=|\mathcal D_{\mathrm{cal}}|\), and let \(\hat q_t^*\) denote the score of a new test trajectory at timestep \(t\). We now define the three procedures.

\paragraph{Fixed-Time Conformal Detection}
Let \(\hat q_{t^\star,(1)} \le \cdots \le \hat q_{t^\star,(n_{\mathrm{cal}})}\) denote the ordered scores of the null calibration trajectories at \(t^\star\). For a target false-alarm level \(\alpha\in(0,1)\), define
\begin{equation}
    k_\alpha^{\mathrm{fix}} = \lceil (n_{\mathrm{cal}}+1)(1-\alpha)\rceil,
    \qquad
    c_{t^\star,\alpha} = \hat q_{t^\star,(k_\alpha^{\mathrm{fix}})}.
    \label{eq:fixed-threshold}
\end{equation}
Thus, \(c_{t^\star,\alpha}\) is the \(k_\alpha^{\mathrm{fix}}\)th smallest calibration score at \(t^\star\). We use the convention \(\hat q_{t^\star,(n_{\mathrm{cal}}+1)}=+\infty\) when necessary. For a new trajectory, the fixed-time detector alarms at \(t^\star\) if \(\hat q_{t^\star}^{*}>c_{t^\star,\alpha}\). Equivalently, define
\begin{equation}
    \tau_{\mathrm{fix}} =
    \begin{cases}
        t^\star, & \hat q_{t^\star}^{*}>c_{t^\star,\alpha},\\
        +\infty, & \text{otherwise}.
    \end{cases}
    \label{eq:fixed-alarm}
\end{equation}

\paragraph{Bonferroni Sequential Calibration}
As a natural way to obtain trajectory-level validity, we calibrate the score at each timestep using significance level \(\alpha/T\). Let \(\hat q_{t,(1)} \le \cdots \le \hat q_{t,(n_{\mathrm{cal}})}\), with \(\hat q_{t,(n_{\mathrm{cal}}+1)}=+\infty\), denote the ordered null calibration scores at timestep \(t\). We define
\begin{equation}
    k_\alpha^{\mathrm{Bon}} = \lceil (n_{\mathrm{cal}}+1)(1-\alpha/T)\rceil,
    \qquad
    c_{t,\alpha/T} = \hat q_{t,(k_\alpha^{\mathrm{Bon}})}.
    \label{eq:bonferroni-threshold}
\end{equation}
The corresponding alarm time is
\begin{equation}
    \tau_{\mathrm{Bon}} = \inf\bigl\{t\in\{1,\ldots,T\}:\hat q_t^{*}>c_{t,\alpha/T}\bigr\}.
    \label{eq:bonferroni-alarm}
\end{equation}

\paragraph{Trajectory-Level Max Calibration}
Trajectory-level calibration treats the event of crossing a common threshold at any timestep as a single trajectory-level statistic. For each null calibration trajectory, define
\begin{equation}
    Z_i = \max_{t=1,\ldots,T}\hat q_t^{(i)},
    \qquad i=1,\ldots,n_{\mathrm{cal}}.
    \label{eq:trajectory-max-score}
\end{equation}
Let \(Z_{(1)}\le\cdots\le Z_{(n_{\mathrm{cal}})}\) denote the ordered trajectory-level scores, with \(Z_{(n_{\mathrm{cal}}+1)}=+\infty\). Define
\begin{equation}
    k_\alpha^{\mathrm{max}} = \lceil (n_{\mathrm{cal}}+1)(1-\alpha)\rceil,
    \qquad c_\alpha = Z_{(k_\alpha^{\mathrm{max}})}.
    \label{eq:max-threshold}
\end{equation}
For a new trajectory, the corresponding alarm time is
\begin{equation}
    \tau_{\max} = \inf\bigl\{t\in\{1,\ldots,T\}:\hat q_t^{*}>c_\alpha\bigr\},
    \label{eq:max-alarm}
\end{equation}
where \(\inf\emptyset=+\infty\). For all three procedures, \(\tau=+\infty\) means no alarm during the trajectory. If a calibration index equals \(n_{\mathrm{cal}}+1\), the corresponding threshold is \(+\infty\), so it cannot be exceeded.

\subsection{Validity}
\label{sec:validity}

For sequential detection, validity means controlling the probability of ever falsely flagging a generalized trajectory over the entire denoising process. This is important because repeated monitoring across timesteps can inflate the overall false-alarm probability without trajectory-level control.

Our validity results assume that, under $H_0$, the calibration trajectories
\(x_{1:T}^{(1)},\ldots,x_{1:T}^{(n_{\mathrm{cal}})}\)
and a new test trajectory $x_{1:T}^{*}$ are exchangeable, and that the scoring function \(g\) is fixed before calibration. We verify this by construction in Appendix~\ref{appendix:exchangeability}.

The validity of all three strategies follows from the quantile lemma \citep{tibshirani2019conformal}, which gives a finite-sample exceedance bound for an exchangeable test score calibrated by an empirical quantile. Fixed-time calibration applies this result to \(\hat q_{t^\star}\) at level
\(\alpha\), while trajectory-level max calibration applies it to \(\max_{t=1,\ldots,T}\hat q_t\) at level \(\alpha\). Bonferroni calibration applies the same result separately to each \(\hat q_t\) at level \(\alpha/T\), followed by a union bound over timesteps. Consequently, all three strategies satisfy
\begin{equation}
    P_{H_0}(\tau\le T)\le\alpha.
\end{equation}
Detailed proofs are provided in Appendix~\ref{appendix:proofs}.

\section{Experiments}

\subsection{Evaluation Metrics}

The guarantee of Section~\ref{sec:validity} controls \(P_{H_0}(\tau\le T)\),
which we estimate by the false positive rate (FPR) on generalized evaluation trajectories. The true positive rate (TPR) on memorized test trajectories estimates the detection power \(P_{H_1}(\tau\le T)\). An FPR well below \(\alpha\) indicates that a procedure leaves part of its false-alarm budget
unspent.

To characterize detection timing under \(H_1\), we define the decision step \(d=t^\star\) for fixed-time detection and \(d=\min(\tau,T)\) for sequential detection. We report \(\bar\sigma_{\mathrm{dec}}=\mathbb{E}_{H_1}[\sigma_d]\), where larger \(\bar\sigma_{\mathrm{dec}}\) indicates
earlier detection. We also report the cumulative detection probability \(P_{H_1}(\tau\le t)\), whose value at \(t=T\) equals the overall detection
power.

\subsection{Power--Timing Tradeoff Across Settings}
\label{sec:power-timing}

\begin{figure}[!htbp]

    \centering
    
\includegraphics[width=0.8\linewidth]{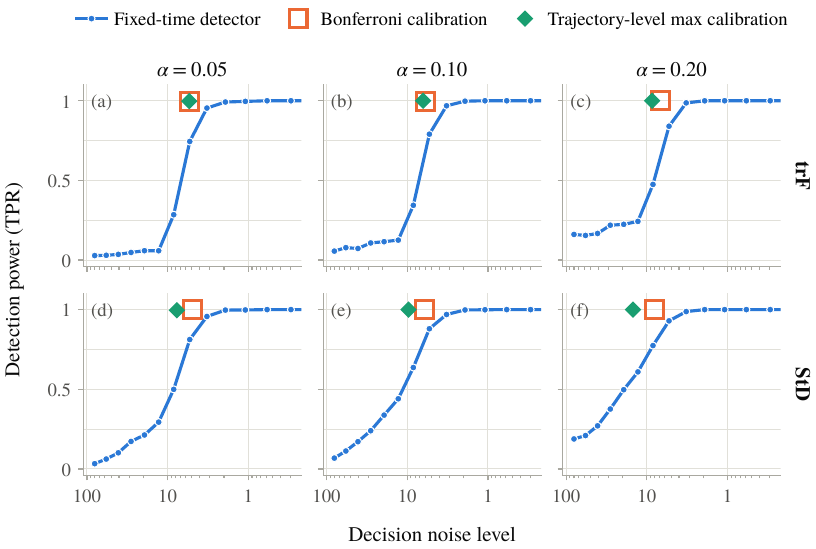}

    \caption{\textbf{Power--timing tradeoff across statistics and false-alarm levels.} Curves show fixed-time detectors across pre-selected noise levels \(\sigma_{t^\star}\); square and diamond markers denote sequential detection under per-step Bonferroni calibration and trajectory-level max calibration, respectively, positioned at their mean decision noise levels \(\bar{\sigma}_{\mathrm{dec}}\) under \(H_1\). The horizontal axis is logarithmic, with higher noise levels corresponding to earlier stages of denoising.
    }
    \label{fig:power-timing-all}
\end{figure}

Figure~\ref{fig:power-timing-all} summarizes the power--timing tradeoff for both statistics and all three false-alarm levels. Fixed-time detectors trace out the operating points obtained by varying the pre-selected noise level \(\sigma_{t^\star}\), while the two sequential calibration strategies are positioned at their respective mean decision noise levels \(\bar{\sigma}_{\mathrm{dec}}\) under \(H_1\).

\FloatBarrier

Across all six settings, both sequential calibration strategies achieve high detection power at relatively high mean decision noise levels, corresponding to earlier stages of denoising. Trajectory-level max calibration is generally shifted further toward higher-noise stages than Bonferroni calibration, particularly for StD. This reflects that trajectories with stronger evidence can be detected early, while harder trajectories continue to be monitored. Consequently, sequential monitoring can attain high detection power while making decisions earlier than fixed-time procedures with comparable power.

\subsection{Cumulative Detection Probability Along the Denoising Trajectory}
\label{subsection:exp1}
\FloatBarrier

\begin{figure}[h]
\centering
\includegraphics[width=0.60\linewidth]{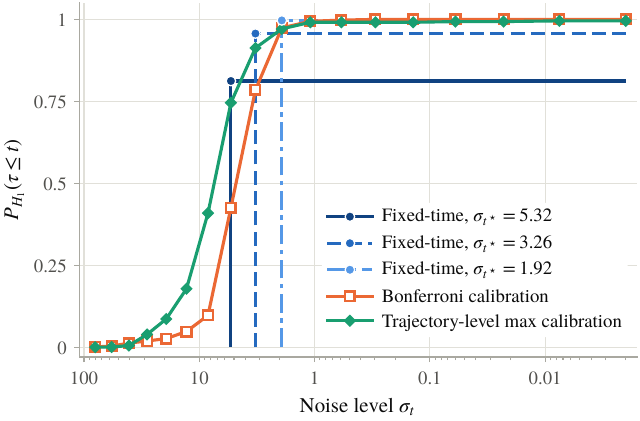}
\caption{\textbf{Cumulative detection probability for StD at \(\alpha=0.05\).}
Fixed-time procedures use pre-selected noise levels \(\sigma\in\{5.32, 3.26, 1.92\}\), corresponding to \(t^\star\in\{8,9,10\}\) for the \(T=18\) sampler; sequential procedures use per-step Bonferroni calibration or trajectory-level max calibration.
}
\label{fig:cumulative-detection}
\end{figure}
\FloatBarrier

Figure~\ref{fig:cumulative-detection} shows how the cumulative detection probability evolves along the denoising trajectory. The fixed-time procedures at \(\sigma\in\{5.32, 3.26, 1.92\}\) are selected to span the transition from moderate detection power to the near-saturated regime. As shown, fixed-time procedures cannot detect a trajectory before its pre-selected noise level is reached, whereas sequential monitoring allows the alarm time to adapt to each trajectory. Among the two sequential calibration strategies, trajectory-level max calibration accumulates detections at higher noise levels than Bonferroni calibration, while both eventually achieve near-perfect detection power.

\subsection{Sequential Calibration Strategies}

\begin{table}[h]
\centering
\small
\caption{\textbf{Comparison of sequential calibration strategies at \(\alpha=0.05\).}
FPR, TPR, and mean decision noise level with standard deviation over five random splits are reported. Larger \(\bar{\sigma}_{\mathrm{dec}}\) corresponds to earlier detection. The empirical FPR may exceed \(\alpha\) on a finite evaluation set.}
\label{tab:sequential-comparison}
\begin{tabular}{llccc}
\toprule
\textbf{Statistic} & \textbf{Calibration} & FPR & TPR & $\bar{\sigma}_{\mathrm{dec}}\;(\uparrow)$ \\
\midrule
\multirow{2}{*}{trF}
& Bonferroni
& $0.021 \pm 0.010$
& $0.994 \pm 0.001$
& $5.329 \pm 0.210$ \\
& Trajectory-level max
& $0.044 \pm 0.008$
& $\mathbf{0.998 \pm 0.000}$
& $\mathbf{5.408 \pm 0.069}$ \\
\midrule
\multirow{2}{*}{StD}
& Bonferroni
& $0.017 \pm 0.008$
& $0.994 \pm 0.003$
& $4.900 \pm 0.336$ \\
& Trajectory-level max
& $0.050 \pm 0.015$
& $\mathbf{0.996 \pm 0.000}$
& $\mathbf{7.678 \pm 0.431}$ \\
\bottomrule
\end{tabular}
\end{table}
Table~\ref{tab:sequential-comparison} compares the two sequential calibration strategies at \(\alpha=0.05\). Bonferroni calibration is more conservative, yielding lower empirical FPRs (\(0.021\) and \(0.017\) for trF and StD): scores at nearby timesteps are strongly dependent, so the per-step crossing events overlap heavily and their probabilities sum to well above the probability of their union. It also requires \(n_{\mathrm{cal}}\ge T/\alpha-1=359\) null calibration trajectories, which our \(n_{\mathrm{cal}}=383\) only narrowly satisfies. Trajectory-level max calibration instead requires only \(n_{\mathrm{cal}}\ge 1/\alpha-1=19\), uses more of the available false-alarm budget, and maintains comparably high detection power. It also yields earlier decisions, particularly for StD, where the mean decision noise level increases from \(4.900\) to \(7.678\). Further discussion of conservativeness and results for \(\alpha\in\{0.10,0.20\}\) are provided in Appendix~\ref{appendix:conservative} and Appendix~\ref{appendix:sequential-additional}.

\FloatBarrier
\section{Conclusion}
\label{sec:conclusion}

We studied prompt-free early detection of memorization as a sequential testing problem. Under exchangeability and with a scoring function fixed before calibration, fixed-time, Bonferroni, and trajectory-level max calibration all provide finite-sample control of the probability of a false alarm. Experiments on an unconditional CIFAR-10 EDM show that sequential monitoring improves the power--timing tradeoff over fixed-time detection. Among the sequential procedures, trajectory-level max calibration uses the false-alarm budget less conservatively than Bonferroni calibration, enabling earlier detection while maintaining comparable detection power. These results show that calibrating the monitoring event at the trajectory level can provide adaptive alarm times and finite-sample false-alarm control.

Our evaluation is limited to an unconditional CIFAR-10 EDM and uses well-separated endpoint similarity thresholds (Section~\ref{sec:memorization}) to define clearly memorized and generalized trajectories. The approach also requires an auxiliary causal scoring model trained on labeled trajectories. Future work should evaluate larger and more heterogeneous models, consider less separated memorization regimes, and investigate how early alarms can be used for inference-time mitigation.
\bibliographystyle{plainnat}   
\bibliography{references}  

\newpage
\appendix
\section{Additional Experimental Setup}

\subsection{Per-step Statistics}
\label{appendix:statistics}
The trF and StD statistics are defined in Section~\ref{sec:statistics-scoring}. For trF (Eq.~\ref{eq:trf-statistic}), \(F_\theta\) is the raw EDM network output and \(u_t\) is its preconditioned network input \citep{karras2022edm}. We estimate the trace using Hutchinson's estimator \citep{hutchinson1990stochastic} with \(K=10\) Rademacher probes. Under EDM preconditioning, this quantity is affinely related, at each fixed noise level, to the score-Jacobian trace, equivalently the Laplacian of the model log-density.

For StD, following \citet{zhang2026balanced}, we use the activation at \texttt{8x8\_block3.norm0} as \(h_t\) in Eq.~\ref{eq:std-statistic}. We apply spatial global-max pooling and compute the channel-wise standard deviation.

\subsection{Scoring Model}
\label{appendix:detector}

For each statistic, we train a separate GRU scoring model. The scoring model uses a one-layer GRU with hidden dimension \(64\), followed by a linear head with sigmoid activation.
The input at each step is the scalar statistic concatenated with the \(16\)-dimensional Fourier time embedding of \(\log \sigma_t\) (\(8\) base frequencies, sine and cosine). We train using AdamW with learning rate \(3\times 10^{-3}\), weight decay \(10^{-4}\), and cosine learning-rate scheduling, batch size \(256\) trajectories, gradient clipping at norm \(1.0\), for at most \(300\) epochs. Early stopping for validation is based on the BCE loss on \(\mathcal{D}_{\mathrm{val}}\) with \(40\) epochs (tolerance \(10^{-5}\)); the weights of the best-validation-loss epoch are retained.

\subsection{Experimental Setup}
\label{appendix:exp-setup}

\paragraph{Model and generation}
The model is the pretrained EDM checkpoint of 
\citet{zhang2026balanced}, an unconditional CIFAR-10 diffusion model trained on the standard \(50000\)-image CIFAR-10 training set augmented with \(100\) duplicated targets (the first ten images per class replicated \(50 \times\) each, adding \(5000\) duplicate samples).

In EDM, \(\sigma\) denotes the noise level conditioning the denoiser: a clean sample is perturbed by Gaussian noise with standard deviation
\(\sigma\), and larger \(\sigma\) therefore corresponds to a noisier, less-denoised state. Sampling proceeds from high to low noise levels \citep{karras2022edm}.
 
We use the deterministic \(18\)-step Heun sampler \citep{karras2022edm} with the standard EDM schedule
\(\sigma_{\min}=0.002\), \(\sigma_{\max}=80.0\), and \(\rho=7\).
The resulting noise levels satisfy
\[
    \sigma_1=80.0>\sigma_2>\cdots>\sigma_T=0.002,
\]
so that step \(1\) corresponds to the initial high-noise state and step \(T\) to the final denoised sample. We generate \(N_{\mathrm{total}}=25000\) trajectories from independent random seeds.

\paragraph{Ground-truth memorization labeling}
Define the training-set similarity score
\begin{equation}
    C(x_T)
    =
    \max_{y \in \mathcal{D}_{\mathrm{train}}}
    \mathrm{SSCD}(x_T,y),
\end{equation}

where $\mathrm{SSCD}(x,y)$ denotes the cosine similarity between the embeddings of $x$ and $y$ produced by a pretrained Self-Supervised Copy
Detection (SSCD) encoder
\citep{pizzi2022sscd}, and
$\mathcal{D}_{\mathrm{train}}$ denotes the training set used to train the diffusion model.  

Following \citet{zhang2026balanced}, we use \(c_{\mathrm{gen}}=0.5\) and \(c_{\mathrm{mem}}=0.9\), and define
\begin{equation}
    L =
    \begin{cases}
        0, & C(x_T)<c_{\mathrm{gen}},\\
        1, & C(x_T)>c_{\mathrm{mem}}.
    \end{cases}
\end{equation}
Trajectories in the intermediate region are excluded. The trajectory distributions used in the main text are then the conditional distributions of \(x_{1:T}\) given \(L=0\) and \(L=1\), denoted by \(P_{\mathrm{gen}}\) and \(P_{\mathrm{mem}}\), respectively.

False-alarm control applies to the \(H_0\) population, while detection power is evaluated on the \(H_1\) population; no guarantee is claimed for intermediate trajectories.

Applying this protocol to our \(25000\) trajectories yields \(2881\) memorized, \(5104\) generalized, and \(17015\) ambiguous.

\paragraph{Data splits}
After removing ambiguous samples, we retain \(7985\) trajectories, including \(2881\) memorized and \(5104\) generalized trajectories.
We split them at the trajectory level into \(\mathcal{D}_{\mathrm{fit}}\), \(\mathcal{D}_{\mathrm{val}}\), and \(\mathcal{D}_{\mathrm{test}}\)
with a \(70\%/15\%/15\%\) ratio, yielding \(5572\), \(1198\), and \(1215\) trajectories, respectively. Memorized trajectories are grouped
by their nearest training example during splitting to prevent the same
duplicated target from appearing across splits.

The GRU is fitted on \(\mathcal{D}_{\mathrm{fit}}\) and early stopping is performed using \(\mathcal{D}_{\mathrm{val}}\). After training, the scoring model is frozen. The test set contains \(449\) memorized and \(766\) generalized trajectories. For each of five random splits, the generalized test trajectories are
divided equally into a null calibration set \(\mathcal{D}_{\mathrm{cal}}\) and a generalized evaluation set, with \(383\) trajectories each; the latter together with all \(449\) memorized test trajectories forms \(\mathcal{D}_{\mathrm{eval}}\). Conformal thresholds are computed only from \(\mathcal{D}_{\mathrm{cal}}\);
FPR is evaluated on the generalized trajectories in \(\mathcal{D}_{\mathrm{eval}}\), while TPR is evaluated on all \(449\) memorized test trajectories.

\paragraph{Compute resources}
Experiments were run on a single NVIDIA RTX 4090 with \(24\) GB GPU memory. A single scoring-model training run takes approximately \(15\) minutes. Generating the \(25000\)-trajectory statistics takes approximately \(6\) GPU-hours for the trF signal and \(2\) GPU-hours for the StD signal; downstream labeling, calibration, and evaluation across the five random splits together take under one additional GPU-hour. The full pipeline for both signals therefore requires approximately \(12\)--\(15\) total GPU-hours. Preliminary experiments during method development (exploring alternative statistics, StD layer choices, scoring architectures, labeling protocols, and calibration variants) required approximately \(30\) additional GPU-hours.

\paragraph{Existing assets}
We use CIFAR-10, the EDM implementation of
\citet{karras2022edm} and the SSCD encoder of \citet{pizzi2022sscd}. The EDM code and pretrained models are released under CC BY-NC-SA 4.0, and the SSCD codebase under the MIT License. CIFAR-10 is used according to its original distribution terms. The pretrained CIFAR-10 EDM checkpoint of \citet{zhang2026balanced} we applied was provided by the original authors and used with their permission.

\subsection{Exchangeability by Construction}
\label{appendix:exchangeability}

The validity results rely on exchangeability of the null calibration and
test trajectories, conditional on \(\mathcal{D}_{\mathrm{fit}}\) and \(\mathcal{D}_{\mathrm{val}}\), which holds by construction for three reasons.

First, all \(25000\) trajectories are drawn from the same frozen EDM checkpoint using the same deterministic sampler and the same noise schedule, with initial noise drawn from independent seeds. The trajectories are therefore i.i.d.\ draws from a single distribution, and i.i.d.\ implies exchangeable.

Second, the same SSCD encoder, the same training-image index, and the same thresholds are applied to every trajectory. Conditioning on \(L=0\) is a selection operator applied identically to all trajectories, so the null pool remains exchangeable; calibration and evaluation trajectories are drawn from this common pool.

Finally, the scoring model \(g\) is fitted on \(\mathcal{D}_{\mathrm{fit}}\) and early-stopped on
\(\mathcal{D}_{\mathrm{val}}\), both disjoint from
\(\mathcal{D}_{\mathrm{test}}\). Conditional on
\(\mathcal{D}_{\mathrm{fit}}\) and \(\mathcal{D}_{\mathrm{val}}\), \(Z=\max_t g(M_{1:t})\) is a fixed measurable map applied identically to every trajectory, and
exchangeability is preserved under a common measurable transformation. The trF statistic additionally depends on the probes. They are drawn i.i.d.\ across trajectories and independently of the sampling seeds, so \(Z\) is a common randomized measurable map and exchangeability is preserved.

\section{Sequential Calibration Across False-Alarm Levels}
\label{appendix:sequential-additional}

\begin{table}[h]
\centering
\small
\caption{\textbf{Comparison of sequential calibration strategies at \(\alpha=0.10\) and \(\alpha=0.20\).}
FPR, TPR, and mean decision noise level with standard deviation over five random splits are reported. Larger \(\bar{\sigma}_{\mathrm{dec}}\) corresponds to earlier detection.}
\label{tab:sequential-additional}
\begin{tabular}{lllccc}
\toprule
$\alpha$ & \textbf{Statistic} & \textbf{Calibration}
& FPR & TPR & $\bar{\sigma}_{\mathrm{dec}}\;(\uparrow)$ \\
\midrule

\multirow{4}{*}{$0.10$}
& \multirow{2}{*}{trF}
  & Bonferroni           & $0.045 \pm 0.010$ & $0.998 \pm 0.002$ & $5.920 \pm 0.337$ \\
& & Trajectory-level max & $0.084 \pm 0.017$ & $\mathbf{1.000 \pm 0.000}$ & $\mathbf{6.337 \pm 0.212}$ \\
\cmidrule(l){2-6}
& \multirow{2}{*}{StD}
  & Bonferroni           & $0.030 \pm 0.002$ & $\mathbf{1.000 \pm 0.000}$ & $6.140 \pm 0.292$ \\
& & Trajectory-level max & $0.091 \pm 0.021$ & $\mathbf{1.000 \pm 0.000}$ & $\mathbf{9.638 \pm 0.463}$ \\
\midrule
\multirow{4}{*}{$0.20$}
& \multirow{2}{*}{trF}
  & Bonferroni           & $0.076 \pm 0.004$ & $\mathbf{1.000 \pm 0.000}$ & $6.692 \pm 0.168$ \\
& & Trajectory-level max & $0.172 \pm 0.014$ & $\mathbf{1.000 \pm 0.000}$ & $\mathbf{8.603 \pm 0.060}$ \\
\cmidrule(l){2-6}
& \multirow{2}{*}{StD}
  & Bonferroni           & $0.072 \pm 0.021$ & $\mathbf{1.000 \pm 0.000}$ & $7.986 \pm 0.864$ \\
& & Trajectory-level max & $0.202 \pm 0.013$ & $\mathbf{1.000 \pm 0.000}$ & $\mathbf{14.862 \pm 0.279}$ \\

\bottomrule
\end{tabular}
\end{table}

Table~\ref{tab:sequential-additional} extends the sequential calibration comparison in Table~\ref{tab:sequential-comparison} to
\(\alpha\in\{0.10,0.20\}\). The same qualitative behavior is observed: per-step Bonferroni calibration yields lower empirical FPRs, whereas trajectory-level max calibration uses more of the available false-alarm budget while maintaining high detection power. Across these settings, trajectory-level calibration also shifts decisions toward higher-noise, less-denoised stages of the reverse process.

\section{Proofs}
\label{appendix:proofs}
Throughout this section we assume the exchangeability condition of Section~\ref{sec:validity}. All alarm times take values in
\(\{1,\ldots,T\}\cup\{+\infty\}\), where \(+\infty\) denotes that no alarm is raised within the finite horizon. All rejection rules use strict inequalities, so the conformal bounds remain valid in the presence of ties.

\subsection{Validity of Fixed-Time Conformal Detection}

For the pre-selected timestep \(t^\star\), exchangeability of the null trajectories and the fact that \(g\) is fixed before calibration imply that
\[
    \hat q_{t^\star}^{(1)},\ldots,
    \hat q_{t^\star}^{(n_{\mathrm{cal}})},
    \hat q_{t^\star}^{*}
\]
are exchangeable.

By the quantile lemma,
\begin{equation}
\begin{aligned}
    P_{H_0}
    \bigl(
        \hat q_{t^\star}^{*} > c_{t^\star,\alpha}
    \bigr)
    &\le
    \frac{n_{\mathrm{cal}}+1-k_\alpha^\mathrm{fix}}{n_{\mathrm{cal}}+1} \\
    &\le
    \alpha .
\end{aligned}
\end{equation}

Since
\[
    \tau_{\mathrm{fix}}
    =
    \begin{cases}
        t^\star,
        & \hat q_{t^\star}^{*}>c_{t^\star,\alpha},\\
        \infty,
        & \text{otherwise},
    \end{cases}
\]
we have
\begin{equation}
    \{\tau_{\mathrm{fix}}\le T\}
    =
    \{
        \hat q_{t^\star}^{*}>c_{t^\star,\alpha}
    \}.
\end{equation}
Hence,
\begin{equation}
        P_{H_0}(\tau_{\mathrm{fix}}\le T)
        \le
        \alpha.
\end{equation}

\subsection{Validity of Bonferroni Sequential Detection}

For any fixed timestep \(t\), exchangeability of the null trajectories and the fact that \(g\) is fixed before calibration imply that
\[
\hat q_t^{(1)},\ldots,
\hat q_t^{(n_{\mathrm{cal}})},\hat q_t^*
\]
are exchangeable. For each timestep \(t\), the quantile lemma gives
\begin{equation}
    \begin{aligned}
    P_{H_0}
    \bigl(
        \hat q_t^* > c_{t,\alpha/T}
    \bigr)
    &\le
    \frac{n_{\mathrm{cal}}+1-k_\alpha^{\mathrm{Bon}}}{n_{\mathrm{cal}}+1} \\
    &\le
    \frac{\alpha}{T} .
\end{aligned}
\end{equation}

Define the threshold-crossing event at timestep \(t\) as
\begin{equation}
    A_t
    =
    \{
        \hat q_t^* > c_{t,\alpha/T}
    \}.
\end{equation}
The Bonferroni alarm time is
\begin{equation}
    \tau_{\mathrm{Bon}}
    =
    \inf
    \bigl\{
        t\in\{1,\ldots,T\} :
        \hat q_t^* > c_{t,\alpha/T}
    \bigr\}.
\end{equation}
The detector alarms by time \(T\) if and only if at least one of the events
\(A_1,\ldots,A_T\) occurs. Hence,
\begin{equation}
\begin{aligned}
    P_{H_0}(\tau_{\mathrm{Bon}} \le T)
    &=
    P_{H_0}
    \biggl(
        \bigcup_{t=1}^{T} A_t
    \biggr) \\
    &\le
    \sum_{t=1}^{T}
    P_{H_0}(A_t) \\
    &\le
    \sum_{t=1}^{T}
    \frac{\alpha}{T} \\
    &=
    \alpha.
\end{aligned}
\end{equation}
Therefore,
\begin{equation}
        P_{H_0}(\tau_{\mathrm{Bon}} \le T)
        \le
        \alpha.
\end{equation}

\subsection{Validity of Trajectory-Level Max Calibration}

Recall that
\begin{equation}
    Z_i
    =
    \max_{t=1,\ldots,T} \hat q_t^{(i)},
    \qquad
    i=1,\ldots,n_{\mathrm{cal}},
\end{equation}
and
\begin{equation}
    Z_*
    =
    \max_{t=1,\ldots,T} \hat q_t^{*}.
\end{equation}
Because the same fixed scoring function and the same maximum operator are applied to every trajectory, exchangeability of the trajectories implies
exchangeability of
\[
Z_1,\ldots,Z_{n_{\mathrm{cal}}},Z_*.
\]

Applying the quantile lemma under exchangeability,
\begin{equation}
\begin{aligned}
    P_{H_0}(Z_* > c_\alpha)
    &\le
    \frac{n_{\mathrm{cal}}+1-k_\alpha^\mathrm{max}}{n_{\mathrm{cal}}+1} \\
    &\le
    \alpha .
\end{aligned}
\end{equation}

Under trajectory-level max calibration, the sequential alarm time is
\begin{equation}
    \tau_{\max}
    =
    \inf
    \bigl\{
        t\in\{1,\ldots,T\} :
        \hat q_t^* > c_\alpha
    \bigr\},
\end{equation}
where \(\inf\emptyset = +\infty\), the event that the detector alarms at any point along the trajectory is
exactly
\begin{equation}
\begin{aligned}
    \{\tau_{\max} \le T\}
    &=
    \{\exists t \le T :
        \hat q_t^* > c_\alpha\} \\
    &=
    \{\max_{t=1,\ldots,T}
        \hat q_t^* > c_\alpha\} \\
    &=
    \{Z_* > c_\alpha\}.
\end{aligned}
\end{equation}
Hence,
\begin{equation}
        P_{H_0}(\tau_{\max} \le T)
        \le
        \alpha.
\end{equation}

\subsection{Comparison of Bonferroni and Trajectory-Level Calibration}
\label{appendix:conservative}

\paragraph{Conservativeness}
The Bonferroni guarantee follows from the union bound,
\begin{equation}
    P_{H_0}
    \biggl(
        \bigcup_{t=1}^{T} A_t
    \biggr)
    \le
    \sum_{t=1}^{T} P_{H_0}(A_t).
\end{equation}

In diffusion trajectories, scores at nearby timesteps can be strongly
dependent, so the events \(A_t\) may overlap substantially. In this case,
the union bound can contain considerable slack, and the realized
trajectory-level false-alarm probability can be substantially below
\(\alpha\).

An alternative is to calibrate the trajectory-level maximum
\[
    \max_{t=1,\ldots,T} \hat q_t
\]
directly under \(H_0\). This treats the entire monitoring event as a single trajectory-level statistic and therefore avoids splitting the false-alarm budget across timesteps. When timestep-wise crossing events are strongly dependent, this calibration can be less conservative than a Bonferroni construction.

\paragraph{Calibration-size requirement}
Beyond conservativeness, the two procedures differ in how much null data they require. For the per-step threshold \(c_{t,\alpha/T}\) to be finite, the order statistic index must satisfy
\(k_\alpha^{\mathrm{Bon}} \le n_{\mathrm{cal}}\), which requires
\begin{equation}
    n_{\mathrm{cal}} \ge \frac{T}{\alpha} - 1 .
\end{equation}
Direct calibration of the trajectory-level maximum instead requires only \(n_{\mathrm{cal}} \ge 1/\alpha - 1\), independent of the horizon \(T\).
At \(\alpha = 0.05\) and \(T = 18\), these bounds are \(359\) and \(19\)
respectively.

\newpage

\end{document}